\documentclass[letterpaper]{article} % DO NOT CHANGE THIS

\usepackage{aaai2027} % DO NOT CHANGE THIS

\nocopyright

\usepackage{amsmath}
\usepackage{graphicx}
\usepackage{amssymb}
\usepackage[hyphens]{url}  % DO NOT CHANGE THIS
\usepackage{natbib}  % DO NOT CHANGE THIS AND DO NOT ADD ANY OPTIONS TO IT
\usepackage{caption} % DO NOT CHANGE THIS AND DO NOT ADD ANY OPTIONS TO IT

\usepackage{algorithm}
\usepackage{algorithmic}

\usepackage{newfloat}
\usepackage{listings}

\DeclareCaptionStyle{ruled}{labelfont=normalfont,labelsep=colon,strut=off}

\floatstyle{ruled}
\newfloat{listing}{tb}{lst}{}
\floatname{listing}{Listing}

\usepackage{booktabs}

\title{CAER: Conflict-Aware Evidence Routing with Dual Prefix Experts for Multimodal Large Language Models}

\author{
    Zixuan Liu,
    Juntao Cai,
    Xiaoxu Cai\thanks{Corresponding author.},
    Haishuai Wang,
    Jiajun Bu
}
\affiliations{
    Zhejiang Key Lab of Accessible Perception and Intelligent Systems\\
    Zhejiang University\\
    Hangzhou 310027, China\\
    \{liuzx068,22651382,xiaoxu.cai,haishuai.wang,bjj\}@zju.edu.cn
}

\begin{document}

\maketitle

\begin{abstract}
Multimodal Large Language Models (MLLMs) have demonstrated remarkable capabilities in multimodal understanding and generation. However, when textual inputs conflict with visual evidence, they still suffer from hallucinations and produce responses inconsistent with visual content. Existing approaches mainly rely on decoding strategies, additional training, verification methods, or prompting techniques, but often lack fine-grained conflict localization and conflict-aware generation. In this work, we propose CAER, a backbone-agnostic framework for visual-language conflict detection and conflict-aware generation. CAER introduces a span-grounded evidence router that transforms claim representations into soft textual queries and retrieves corresponding evidence from frozen visual tokens, enabling fine-grained conflict estimation. Furthermore, we design a dual-prefix expert routing mechanism that learns separate experts for visually supported and contradicted inputs, enabling conflict-aware generation through explicit expert selection. Experiments on the public MMMC benchmark and our newly curated AgriConflict dataset demonstrate that CAER effectively detects visual-language conflicts and improves the reliability of open-source MLLMs without updating their backbone parameters.
\end{abstract}


\section{Introduction}

Multimodal Large Language Models (MLLMs) have recently advanced from visual perception systems toward general-purpose multimodal reasoning models, demonstrating remarkable capabilities in visual question answering, image understanding, and complex visual reasoning tasks \cite{liu2023visual,bai2023qwenvl}. Benefiting from large-scale vision-language pretraining and instruction tuning, modern MLLMs are capable of integrating visual information with linguistic knowledge to solve increasingly challenging multimodal tasks \cite{zhang2025crossmodal}. However, the reliability of MLLMs remains a critical concern, as they frequently generate responses that are inconsistent with the observed visual content, a phenomenon commonly referred to as multimodal hallucination \cite{leng2025curse,dong2025mirage,bai2024hallucination,tu2025ode}.

\begin{figure}[t]
    \centering
    \includegraphics[width=\linewidth]{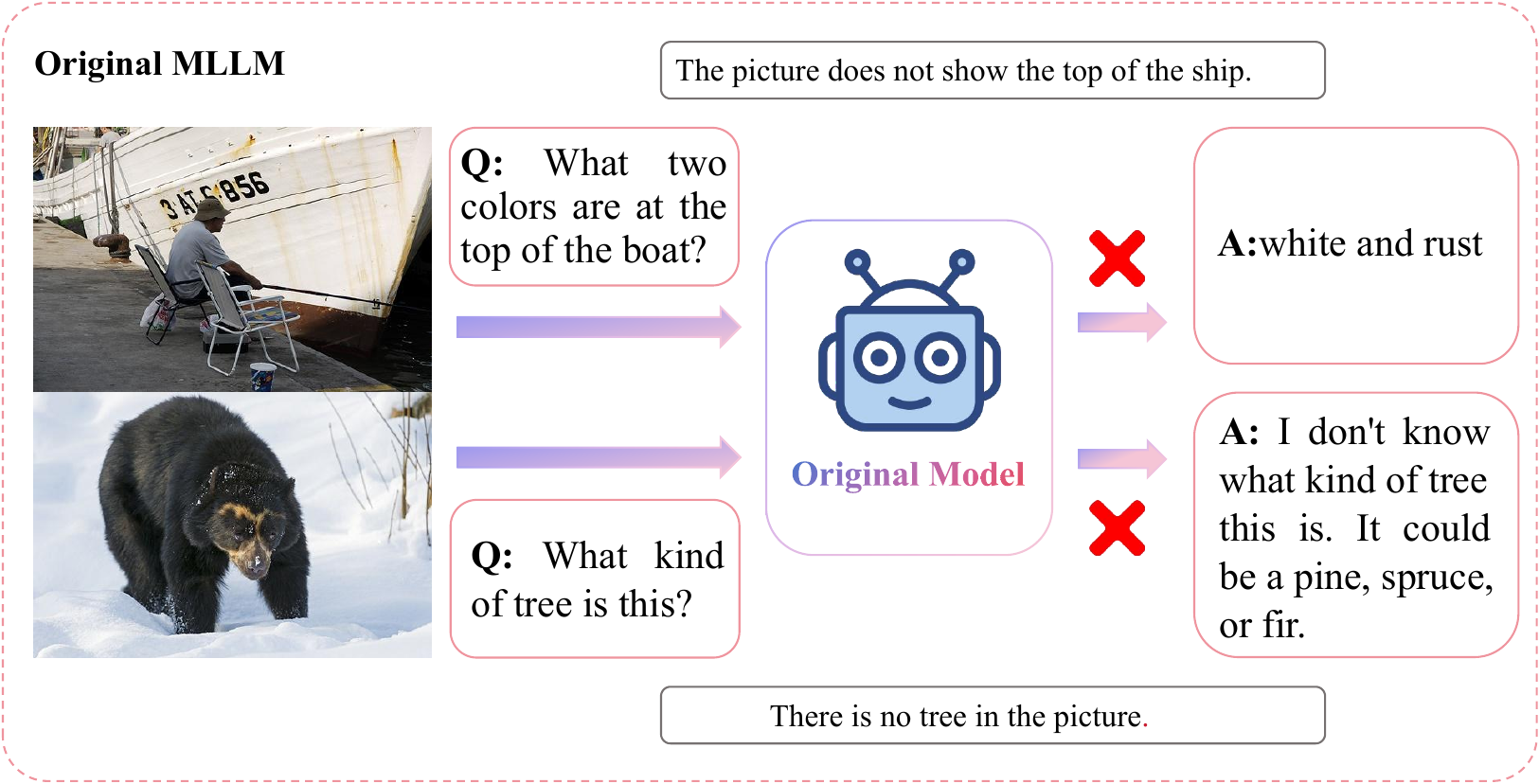}
    \caption{Examples of vision-language conflicts where textual inputs contain false presuppositions inconsistent with visual evidence. MLLM tend to follow the misleading textual assumptions and produce hallucinated responses instead of grounding their answers on the image content.}
    \label{fig:conflict_example}
\end{figure}

Recent studies indicate that hallucination is caused not only by insufficient visual perception, but also by ineffective coordination between visual evidence and linguistic representations \cite{dong2025mirage,li2026unleashing,liu2026adaptive}. As illustrated in Figure~\ref{fig:conflict_example}, MLLMs may rely on semantic patterns inherited from pretrained language models and generate plausible responses despite insufficient or contradictory visual evidence \cite{fang2025grounding,leng2025curse,yin2025clearsight,wu2025mitigating}. This issue becomes more challenging when input texts contain assumptions conflicting with the image. In such cases, models should verify whether textual claims are supported by visual evidence before generation \cite{min2026perception}. Nevertheless, existing MLLMs often follow misleading textual cues rather than visual observations, resulting in unreliable outputs. Recent studies further show that multimodal conflicts are an important yet underexplored source of hallucination in MLLMs \cite{zhang2025crossmodal}.

Existing efforts have improved the reliability of MLLMs from several perspectives. Inference-time approaches enhance visual grounding, such as ClearSight, which strengthens visual representations and reduces language prior dominance without additional training \cite{yin2025clearsight}, and INTER, which introduces interaction guidance to exploit multimodal relationships \cite{dong2025inter}. However, these methods mainly optimize generation behaviors without identifying conflict-related textual claims. Training-based approaches, such as multimodal preference optimization, construct preference data to encourage visually grounded responses \cite{wu2025mitigating}, but require updating MLLM parameters and optimize responses rather than localized conflicts. More recently, modality conflict has been studied as a specific failure mode of MLLMs. Zhang \cite{zhang2025robust} introduce the MMMC benchmark and explore prompting, supervised fine-tuning, and reinforcement learning strategies to improve robustness under conflicting inputs. Nevertheless, existing methods still lack fine-grained conflict localization and evidence-aware generation.

To address these challenges, we propose Conflict-Aware Evidence Routing with Dual Prefix Experts (CAER), a lightweight and backbone-agnostic framework for visual-language conflict detection and conflict-aware generation. Given a textual query containing a false presupposition, CAER identifies conflicting spans and retrieves corresponding visual evidence from frozen representations. Based on textual-visual consistency, CAER activates specialized experts to generate responses aligned with visual observations.

Specifically, CAER adopts a two-stage architecture. First, we introduce a span-grounded evidence router that transforms claim representations into soft textual queries and retrieves evidence from frozen visual tokens, establishing fine-grained correspondence for conflict estimation. Second, we propose a dual-prefix expert routing mechanism that learns separate experts for supported and contradicted inputs. Through explicit expert selection, CAER enables conflict-aware generation without modifying the MLLM backbone, supporting efficient adaptation across architectures while preserving multimodal capabilities.

Our contributions are summarized as follows:

\begin{itemize}

\item We propose a span-grounded evidence routing mechanism that performs claim-level conflict localization by associating textual claims with visual evidence, enabling explicit visual-language consistency estimation.

\item We design a dual-prefix expert routing mechanism that enables conflict-aware generation without updating the underlying MLLM backbone. By learning separate experts for supported and contradicted inputs, CAER adapts generation behavior while preserving multimodal capabilities.

\item We construct AgriConflict, a newly curated dataset for evaluating visual-language conflict understanding in agricultural scenarios. Experiments on MMMC and AgriConflict demonstrate that CAER improves conflict detection and conflict-aware generation across multiple MLLM architectures.

\end{itemize}

\section{Related Works}

\subsection{Vision-Language Conflict Detection}

Early studies on vision-language consistency mainly investigate whether textual statements are supported by visual observations. Visual entailment formulates this problem as determining the relationship between an image and a textual hypothesis, including entailment, contradiction, and neutral relations \cite{xie2019visual}. Subsequent studies extend visual entailment toward fine-grained reasoning by identifying consistency at the semantic element level \cite{thomas2022fine}. These methods establish benchmarks for measuring image-text consistency, but mainly focus on predicting whether statements agree with visual content rather than enabling MLLMs to detect conflicts and adapt responses during generation.

With the emergence of large MLLMs, recent works reveal that conflicts between visual observations and textual information are an important reliability issue. Zhang \cite{zhang2025robust} introduce the MMMC benchmark to evaluate multimodal modality conflict, where textual inputs contain information inconsistent with visual evidence. Their analysis shows that MLLMs may follow misleading textual cues instead of verifying them against images. However, existing benchmarks mainly evaluate robustness under conflicting inputs without providing fine-grained conflict localization.

Recent studies analyze how MLLMs behave under conflicting modalities. Golovanevsky \cite{golovanevsky2025pixels} investigate the competition between visual evidence and pretrained knowledge priors through visual counterfactuals, while Deng \cite{deng2025words} reveal that VLMs may blindly follow textual information under inconsistent inputs. Although these studies provide insights into modality conflicts, they mainly analyze model behaviors rather than enabling explicit conflict detection and conflict-aware generation. Recent works on grounded multimodal alignment further explore fine-grained correspondence between language and visual evidence. Le \cite{le2025progressive} propose progressive multi-granular vision-language alignment to improve grounded reasoning through cross-modal alignment at different semantic levels. However, these methods focus on general multimodal understanding and do not address cases where textual claims contradict visual evidence.

\subsection{Conflict Resolution for Multimodal Models}

Existing approaches for resolving vision-language conflicts can be divided into alignment-based and adaptation-based methods. Alignment-based approaches reduce modality inconsistency by improving interactions between visual and textual representations. Li \cite{li2026deepalign} propose DeepAlign to mitigate modality conflicts through modality-specific alignment, encouraging models to preserve complementary information across modalities. Uni-X \cite{hao2026unix} further explores architectural separation to alleviate conflicts caused by shared representations. Although these methods improve global modality compatibility, they mainly modify representation learning and require additional training, making them unable to explicitly determine whether individual textual claims are supported or contradicted by visual evidence during inference.

Another direction adapts models through additional optimization objectives or reasoning strategies. Zhang \cite{zhang2025robust} evaluate supervised fine-tuning and reinforcement learning strategies on MMMC, demonstrating that post-training improves robustness against modality conflicts. Other approaches introduce specialized objectives to encourage better multimodal reasoning under challenging inputs. However, these methods generally optimize model behavior globally and lack mechanisms for dynamically selecting generation strategies based on detected conflict states.

In contrast, CAER focuses on conflict-aware generation through fine-grained evidence routing. Instead of modifying the entire multimodal model or relying on global alignment objectives, CAER explicitly associates textual claims with visual evidence and selects specialized generation experts according to conflict states while keeping the original MLLM backbone frozen.

\section{Motivation Analysis}
Although recent open-source MLLMs have achieved remarkable progress in multimodal understanding and reasoning, their ability to handle cases where textual inputs contain false presuppositions that conflict with visual evidence remains underexplored. To investigate this limitation, we conduct a preliminary analysis of representative open-source MLLMs on vision-language conflict scenarios. Specifically, we analyze their response behaviors under conflicting inputs, including whether they can identify false presuppositions, rely on visual evidence, and maintain consistent responses across paired conflict and non-conflict settings.

\subsection{Evaluation Setup}

\textbf{Dataset.}
We conduct our analysis on the updated MMMC benchmark \cite{zhang2025robust}, which contains 40,000 image-question samples. Following the official split, we use the 4,000-sample test set for evaluation. To improve textual quality, we employ Qwen3-1.7B to identify issues such as grammatical errors and meaningless symbols, followed by GPT-5.4-mini \cite{openai2026gpt54} for revision. Unlike conventional VQA datasets, MMMC evaluates multimodal models under vision-language conflicts, where textual inputs may contain premises inconsistent with visual content. Each image is associated with paired questions, including a conflict question with an inconsistent premise and a corresponding non-conflict question supported by visual evidence, enabling analysis of conflict resolution and response consistency.

\textbf{Models.}
We evaluate six representative open-source MLLMs with publicly available checkpoints, including InternVL3-8B \cite{zhu2025internvl3}, LLaVA-OV1.5-8B \cite{an2025llavaonevision15}, LLaVA-OV2-8B, Ministral-3-8B \cite{liu2026ministral}, Qwen3-VL-8B \cite{bai2025qwen3vl}, and Qwen3.5-9B. All models are evaluated under the zero-shot setting without conflict-specific adaptation.

\textbf{Evaluation Protocol.}
Since conflict resolution requires semantic understanding beyond exact matching, we employ GPT-5.4-mini \cite{openai2026gpt54} as an automatic evaluator. The evaluator receives the image, question, reference answer, and model response, and categorizes responses according to predefined behaviors.

For conflict questions, we define four categories. Correct Premise Correction Accuracy (CRA) measures cases where the model identifies the false presupposition and generates responses consistent with visual evidence. False Premise Hallucination Rate (HAR) measures cases where the model accepts the false premise and responds based on the incorrect assumption. Prior-Guided Response Rate (PGR) measures cases where the model relies on prior knowledge or unsupported guesses rather than visual evidence. Evasion Rate (ER) measures cases where the model refuses to answer or produces irrelevant responses.

For non-conflict questions, we measure Normal Correct Answer Rate (NCA) and Over-Refusal Rate (ORR) to evaluate whether models preserve normal visual understanding without unnecessary rejection. Based on the paired design of MMMC, we further introduce two image-level consistency measures: Preserved Success Rate (PSR), measuring images where the model correctly answers both the non-conflict and corresponding conflict questions, and Presupposition-Induced Failure Rate (PIFR), measuring images where the model answers the non-conflict question correctly but fails on the conflict question due to accepting the false premise.

\subsection{Observations on Existing MLLMs}

\textbf{Observation 1: Existing MLLMs frequently accept conflicting textual assumptions instead of correcting them based on visual evidence.}

Table~\ref{tab:conflict_results} summarizes response behaviors of different MLLMs under conflict scenarios. Most evaluated models exhibit substantial false premise hallucinations. For example, InternVL3-8B achieves only 22.40\% CRA while producing false premise hallucinations in 73.95\% of conflict cases. Similarly, LLaVA-OV2-8B obtains 18.40\% CRA and 68.80\% HAR. These results indicate that current MLLMs struggle to verify textual assumptions against visual evidence and tend to preserve provided premises during generation.

\begin{table}[t]
\centering
\caption{Response behaviors of open-source MLLMs on conflict questions.}
\label{tab:conflict_results}
\resizebox{\linewidth}{!}{
\begin{tabular}{lcccc}
\toprule
Model & CRA$\uparrow$ & HAR$\downarrow$ & PGR$\downarrow$ & ER$\downarrow$ \\
\midrule
InternVL3-8B & 22.40 & 73.95 & 2.40 & 0.60 \\
LLaVA-OV1.5-8B & 24.60 & 57.75 & 11.25 & 3.40 \\
LLaVA-OV2-8B & 18.40 & 68.80 & 7.25 & 2.90 \\
Ministral-3-8B & 59.00 & 34.65 & 3.05 & 1.50 \\
Qwen3-VL-8B & 63.40 & 32.55 & 2.85 & 0.40 \\
Qwen3.5-9B & 62.35 & 33.50 & 2.70 & 0.35 \\
\bottomrule
\end{tabular}
}
\end{table}

\textbf{Observation 2: Stronger MLLMs improve conflict handling, but still suffer from unresolved conflicts.}

Compared with earlier models, Qwen3-VL-8B and Qwen3.5-9B achieve higher CRA values of 63.40\% and 62.35\%, respectively. However, both models still produce false premise hallucinations in over 30\% of conflict cases. This indicates that stronger multimodal capabilities improve conflict handling but do not inherently enable identifying and resolving inconsistent textual claims.

\textbf{Observation 3: Conflict failures are not mainly caused by insufficient visual understanding, but by inconsistent utilization of visual evidence.}

The paired evaluation results in Table~\ref{tab:normal_results} provide further insights. Existing models generally achieve reasonable performance on non-conflict questions, with NCA ranging from 57.45\% to 74.30\%. However, conflicting textual premises on the same images can substantially degrade performance. For instance, InternVL3-8B correctly answers non-conflict questions but produces presupposition-induced failures on 52.25\% of corresponding images. This gap indicates that models can extract relevant visual information but lack the ability to determine when textual inputs should be trusted or corrected.

\begin{table}[t]
\centering
\caption{Performance of open-source MLLMs on non-conflict questions.}
\label{tab:normal_results}
\resizebox{\linewidth}{!}{
\begin{tabular}{lcccc}
\toprule
Model & NCA$\uparrow$ & ORR$\downarrow$ & PSR$\uparrow$ & PIFR$\downarrow$ \\
\midrule
InternVL3-8B & 69.70 & 0.50 & 15.15 & 52.25 \\
LLaVA-OV1.5-8B & 69.95 & 0.70 & 18.15 & 40.70 \\
LLaVA-OV2-8B & 65.20 & 1.10 & 12.25 & 45.45 \\
Ministral-3-8B & 57.45 & 5.70 & 33.65 & 21.35 \\
Qwen3-VL-8B & 72.50 & 1.25 & 46.45 & 23.40 \\
Qwen3.5-9B & 74.30 & 0.85 & 47.25 & 24.35 \\
\bottomrule
\end{tabular}
}
\end{table}

The above analysis reveals two limitations of current MLLMs. First, models lack fine-grained mechanisms to associate conflicting textual claims with corresponding visual evidence. Second, models do not adapt their generation behavior according to the detected conflict state: they either follow misleading textual assumptions or rely on unnecessary refusal behaviors. These observations motivate the design of CAER, which explicitly performs evidence-grounded conflict routing and selects specialized generation experts according to the estimated conflict state.

\section{Method}

\subsection{Framework Formulation}

Given an image--text pair \((I,x)\), CAER aims to determine whether the textual claim is supported by visual evidence or contains a conflict with the observed image, and subsequently generate a response conditioned on the detected conflict status. We define the conflict status label as \(y\in\{0,1\}\), where \(y=0\) represents visually supported inputs and \(y=1\) represents conflicting inputs requiring correction.

\begin{figure*}[t]
    \centering
    \includegraphics[width=\textwidth]{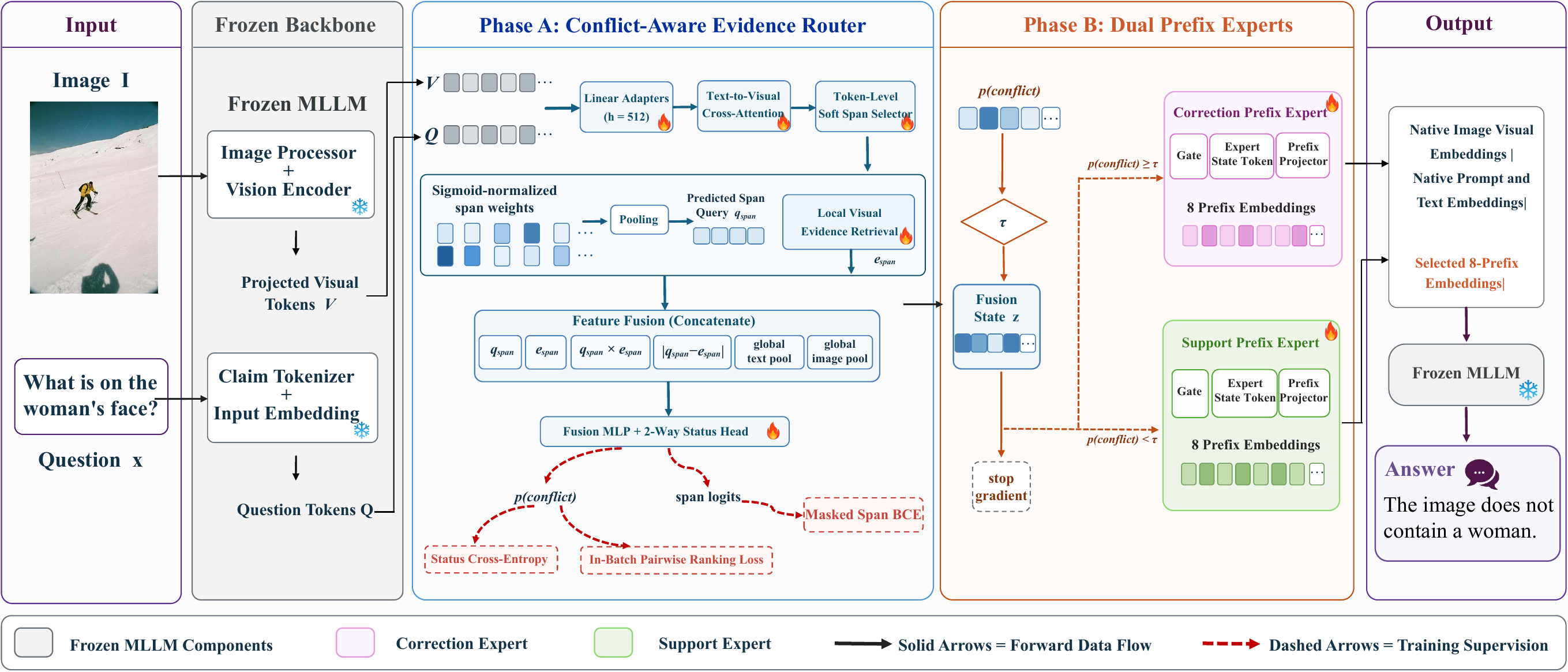}
    \caption{Overview of CAER. The framework consists of a span-grounded evidence router and dual prefix experts for conflict-aware generation. The frozen vision-language backbone remains unchanged, while the proposed modules perform evidence-grounded conflict detection and status-conditioned response generation.}
    \label{fig:framework}
\end{figure*}

As illustrated in Figure~\ref{fig:framework}, CAER consists of a trainable Conflict-Aware Evidence Router and two status-specialized prefix experts built upon a frozen vision-language backbone. Given an input image, the frozen backbone extracts projected visual tokens \(V\in\mathbb{R}^{N_v\times d}\), while the textual claim is tokenized and mapped into language-space embeddings \(Q\in\mathbb{R}^{N_q\times d}\). Different from conventional image-text matching approaches that model conflict as a global similarity problem, CAER performs claim-level conflict reasoning by first identifying conflict-relevant textual spans and then retrieving the corresponding visual evidence required for verification.

CAER is optimized in two stages. In Phase A, we train the evidence router while keeping all parameters of the underlying vision-language backbone frozen. The router estimates the conflict probability \(p_{\mathrm{conflict}}\) and produces an evidence-aware routing representation for subsequent generation. In Phase B, we initialize from the learned router and introduce two independent prefix experts: a support expert for visually supported inputs and a correction expert for conflicting inputs. During inference, CAER performs deterministic binary routing with a predefined threshold \(\tau\):

\begin{equation}
P =
\left\{
\begin{array}{ll}
P_s, & p_{\mathrm{conflict}} < \tau,\\
P_c, & p_{\mathrm{conflict}} \geq \tau .
\end{array}
\right.
\end{equation}

where \(P_s\) and \(P_c\) denote the support and correction prefixes, respectively. The threshold \(\tau\) is determined without accessing test labels. The selected prefix is then used to condition the frozen language decoder for autoregressive generation.

\subsection{Span-Grounded Evidence Router}

The Phase-A router aims to estimate visual-language consistency by grounding conflict detection at the claim level. Specifically, we first project visual tokens \(V\) and textual representations \(Q\) into a shared hidden space. A text-to-visual cross-attention module is then applied to obtain visually conditioned textual representations \(Q_v\), allowing each textual token to incorporate relevant visual information.

Based on these representations, a token-level span selector predicts the relevance of each textual token to potential conflicts. After sigmoid activation, masking, and normalization, the obtained soft weights \(a_i\) are used to aggregate conflict-relevant textual information:

\begin{equation}
q_{\mathrm{span}}
=
\operatorname{LayerNorm}
\left(
\sum_{i=1}^{N_q} a_i Q_i
\right).
\end{equation}

The resulting span query subsequently attends to the visual tokens to retrieve localized visual evidence:

\begin{equation}
e_{\mathrm{span}}
=
\operatorname{CrossAttn}
(q_{\mathrm{span}},V).
\end{equation}

This two-stage grounding process enables CAER to first identify semantically critical textual spans and then retrieve the visual evidence required to verify their consistency with the image.

To capture complementary information from local and global modalities, we combine claim-level evidence with global contextual features. Specifically, we first construct a joint representation:

\begin{equation}
h=
[
q_{\mathrm{span}};
e_{\mathrm{span}};
q_{\mathrm{span}}\odot e_{\mathrm{span}};
|q_{\mathrm{span}}-e_{\mathrm{span}}|;
\overline{Q_v};
\overline{V}
],
\end{equation}

and obtain the routing representation through an MLP:

\begin{equation}
z=\operatorname{MLP}(h).
\end{equation}

Here, \(\overline{Q_v}\) and \(\overline{V}\) denote the global textual and visual summaries, respectively. A two-class status classifier then maps \(z\) into support and correction logits:

\begin{equation}
[l_{\mathrm{support}},l_{\mathrm{correction}}]
=
f_{\mathrm{status}}(z).
\end{equation}

and the conflict probability is computed as:

\begin{equation}
p_{\mathrm{conflict}}
=
\operatorname{softmax}
([l_{\mathrm{support}},l_{\mathrm{correction}}])_1 .
\end{equation}

The router is optimized using status supervision and claim-level span supervision:

\begin{equation}
\mathcal{L}_{A}
=
\mathcal{L}_{\mathrm{status}}
+
\lambda_{1}\mathcal{L}_{\mathrm{span}}
+
\lambda_{2}\mathcal{L}_{\mathrm{rank}} .
\end{equation}

Here, \(\mathcal{L}_{\mathrm{status}}\) is a binary cross-entropy loss for conflict classification. \(\mathcal{L}_{\mathrm{span}}\) is a masked binary cross-entropy loss applied only to samples with reliable claim-span annotations. Samples without reliable span alignment still contribute to status learning. The ranking objective \(\mathcal{L}_{\mathrm{rank}}\) encourages conflict samples to obtain higher correction scores than visually supported samples by increasing the margin between their conflict logits. During inference, CAER relies solely on predicted soft spans and does not require access to external span annotations.

\subsection{Dual Prefix Expert Routing}

After learning the evidence-aware routing representation in Phase A, Phase B converts the estimated conflict state into status-conditioned generation behavior. Starting from the Phase-A checkpoint, we preserve the frozen vision-language backbone and learned router while introducing two independent prefix experts.

Given the routing representation \(z\), each expert generates a status-specific prefix representation:

\begin{equation}
\begin{aligned}
g_e &= \sigma(W_e\,\operatorname{sg}(z)),\\
C_e &= [g_e\operatorname{sg}(z);c_e],
\quad e\in\{s,c\}.
\end{aligned}
\end{equation}

where \(s\) and \(c\) denote the support and correction experts, respectively. \(\operatorname{sg}(\cdot)\) denotes the stop-gradient operation, and \(c_e\) represents a learnable expert-specific token. Separate prefix projectors are then applied to transform \(C_s\) and \(C_c\) into support and correction prefix embeddings, respectively. By detaching the routing representation from the expert optimization process, the generation loss cannot alter the learned conflict-routing behavior through the expert branch, thereby preserving the routing capability acquired in Phase A.

During training, the ground-truth conflict status determines the activated expert. Conflict samples optimize the correction expert, while visually supported samples optimize the support expert. The selected prefix is inserted between the native visual embeddings and textual embeddings.

The frozen MLLM backbone remains unchanged, while the expert-conditioned generation loss optimizes the trainable prefix experts. For visually supported samples, additional preservation and improvement objectives are introduced to prevent performance degradation relative to the original backbone and encourage beneficial expert conditioning. Phase-B optimization only updates prefix expert parameters and does not re-optimize the status classification, span localization, or ranking objectives.

During inference, CAER first predicts the conflict probability through the Phase-A router and selects one prefix expert according to the binary routing rule. The selected expert then conditions the frozen decoder for response generation, enabling conflict-aware generation while maintaining the original capabilities of the pretrained MLLM.

\section{Experiments}

\subsection{Experiment Setup}

We evaluate CAER on two vision-language conflict datasets: the processed MMMC benchmark and our newly curated AgriConflict dataset. For MMMC, we preserve the original 8:1:1 training, validation, and test split. For AgriConflict, we follow the same split protocol with case-level separation to avoid image overlap.

During inference, CAER predicts a conflict probability \(p_{\mathrm{conflict}}\) and routes each input to either the support or correction prefix expert. The routing threshold \(\tau\) is determined without test labels. For MMMC, we set \(\tau=0.5\) according to its paired conflict/non-conflict formulation. For AgriConflict, \(\tau\) is selected based on the best validation accuracy and applied to the test set.

For MMMC, we evaluate responses using GPT-5.4-mini \cite{openai2026gpt54} as an automatic evaluator and report CRA and NCA following our preliminary analysis. For AgriConflict, the output is restricted to binary responses and evaluated by classification accuracy.

To evaluate conflict handling in a domain-specific scenario, we curate AgriConflict from AgMMU \cite{gauba2025agmmu}, an agricultural multimodal understanding benchmark. Unlike the original visual question answering task, our objective is to determine whether a textual claim is consistent with visual evidence. We therefore construct image--text pairs from original images and answers while excluding the original questions.

We perform a two-stage screening process to identify image-answer pairs with visual consistency or conflict. First, Qwen3-VL-2B \cite{bai2025qwen3vl} performs coarse screening of candidate samples, while InternVL2.5-8B \cite{chen2024expanding} conducts secondary verification for uncertain cases. These models are only used for dataset construction and are independent of evaluated models. After filtering, image-answer pairs are converted into binary consistency samples, where each claim is labeled as visually supported or conflicting. The dataset is balanced with equal numbers of supported and conflicting samples. The resulting AgriConflict dataset contains 3,948 image--text pairs. During training and inference, the model receives the following instruction:

\begin{quote}
Determine whether the following text is consistent with the image. Reply with only ``yes'' or ``no''. Do not explain.
\end{quote}

Here, ``yes'' denotes a visually supported claim and ``no'' denotes a conflicting claim. For span supervision, we extract visually relevant spans using GLiNER \cite{zaratiana2024gliner}. When GLiNER fails to provide reliable spans, we apply an InternVL-based fallback strategy and retain only phrases matched verbatim with the original claim.

\subsection{Implementation Details}

All backbone parameters in CAER remain frozen during training. The evidence router adopts a hidden dimension of 512 with eight attention heads and two cross-attention stages for visual conditioning and local evidence retrieval. Claim inputs are truncated to a maximum length of 128 tokens.

In Phase A, we optimize the router for five epochs using AdamW with a learning rate of \(1\times10^{-4}\), weight decay of \(0.01\), a 5\% linear warm-up schedule, and gradient clipping with a maximum norm of 1.0. We use a micro-batch size of 2 with gradient accumulation to obtain an effective batch size of 32. Phase-A mini-batches are status-balanced and independently sampled to support the ranking objective.

In Phase B, we initialize from the best Phase-A checkpoint and train the two prefix experts for one epoch. Each expert generates eight continuous prefix embeddings. We use AdamW with a learning rate of \(5\times10^{-5}\), weight decay of \(0.01\), a 5\% warm-up ratio, and an effective batch size of 32. The preservation and improvement objectives are weighted by 0.5 and 0.25, respectively, and each prefix expert is regularized with an \(\ell_2\) penalty of 0.01.

All experiments use greedy decoding. For AgriConflict, generation is restricted to binary ``yes'' or ``no'' responses.

\subsection{Baselines}

We evaluate CAER with several representative open-source MLLM backbones, including InternVL3-8B \cite{zhu2025internvl3}, LLaVA-OV1.5-8B \cite{an2025llavaonevision15}, LLaVA-OV2-8B, Ministral-3-8B \cite{liu2026ministral}, Qwen3-VL-8B \cite{bai2025qwen3vl}, and Qwen3.5-9B. These models cover diverse vision encoders and language backbones. For each backbone, we report the original model performance as the base reference.

We further compare CAER with representative conflict-handling approaches. Specifically, we include two training-free prompting methods, FoV prompt \cite{liu2025insight} and MMMC prompt \cite{zhang2025robust}, which encourage visual evidence prioritization and textual premise verification, respectively. We also include a supervised adaptation baseline that applies LoRA \cite{hu2022lora} fine-tuning to all linear layers with answer-generation supervision. In addition, we compare with ASCD \cite{wang2026ascd}, an attention-steerable contrastive decoding method that reduces visually ungrounded generation by adjusting cross-modal attention during decoding. These baselines cover prompting, parameter-updating, and decoding-time strategies for multimodal reliability.

\subsection{Main Results}

We compare CAER with original MLLMs, prompting-based approaches, supervised adaptation, and decoding-based methods on MMMC and AgriConflict. The results on MMMC are summarized in Table~\ref{tab:mmmc_cra_main} and Table~\ref{tab:mmmc_nca_main}, while the results on AgriConflict are reported in Table~\ref{tab:agri_conflict_main} and Table~\ref{tab:agri_normal_main}.

\begin{table}[t]
\centering
\caption{CRA (\%) on conflict questions of MMMC.}
\label{tab:mmmc_cra_main}
\resizebox{\linewidth}{!}{
\begin{tabular}{lcccccc}
\toprule
Method & InternVL3 & LLaVA-OV2 & LLaVA-OV1.5 & Ministral3 & Qwen3-VL & Qwen3.5 \\
\midrule
Raw &22.40&18.40&24.60&59.00&63.40&62.35\\
FoV prompt&25.50&7.45&15.35&52.00&64.20&62.50\\
MMMC prompt&46.10&41.45&68.10&65.45&71.50&67.35\\
ASCD&25.25&18.25&27.55&53.05&64.35&61.00\\
LoRA&86.20&85.10&85.50&91.05&85.10&85.65\\
CAER(ours)&82.50&81.40&81.95&81.30&81.25&81.15\\
\bottomrule
\end{tabular}}
\end{table}

\begin{table}[t]
\centering
\caption{NCA accuracy (\%) on non-conflict questions of MMMC.}
\label{tab:mmmc_nca_main}
\resizebox{\linewidth}{!}{
\begin{tabular}{lcccccc}
\toprule
Method & InternVL3 & LLaVA-OV2 & LLaVA-OV1.5 & Ministral3 & Qwen3-VL & Qwen3.5 \\
\midrule
Raw&69.70&65.20&69.95&57.45&72.50&74.30\\
FoV prompt&70.65&68.05&70.40&61.60&74.30&75.65\\
MMMC prompt&64.95&52.50&61.45&52.55&67.80&72.90\\
ASCD&68.65&66.55&70.55&61.25&70.65&73.85\\
LoRA&62.60&61.60&62.55&53.80&66.15&65.45\\
CAER(ours)&58.25&54.70&56.25&52.10&59.90&55.85\\
\bottomrule
\end{tabular}}
\end{table}

CAER consistently improves conflict resolution compared with the corresponding original MLLMs across different backbone architectures. For example, on InternVL3-8B, CAER improves CRA from 22.40\% to 78.35\%, while on LLaVA-OV1.5-8B it increases CRA from 24.60\% to 81.95\%. These improvements demonstrate that explicitly modeling claim-level evidence and introducing conflict-aware routing can effectively reduce the tendency of MLLMs to follow misleading textual premises.

Compared with prompting-based approaches, CAER provides more consistent improvements across different architectures. Although methods such as FoV prompt and MMMC prompt can encourage models to consider visual information, they rely on the original reasoning process of the backbone model and do not explicitly identify conflict-related evidence. In contrast, CAER first estimates visual-language consistency through the span-grounded evidence router and then activates a specialized prefix expert according to the predicted conflict state.

\begin{table}[t]
\centering
\caption{Conflict classification accuracy (\%) on AgriConflict.}
\label{tab:agri_conflict_main}
\resizebox{\linewidth}{!}{
\begin{tabular}{lcccccc}
\toprule
Method & InternVL3 & LLaVA-OV2 & LLaVA-OV1.5 & Ministral3 & Qwen3-VL & Qwen3.5\\
\midrule
Raw & 69.54 & 52.79 & 52.79 & 89.85 & 61.42 & 75.63\\
FoV prompt & 44.16 & 46.70 & 53.30 & 78.17 & 55.84 & 56.85\\
MMMC prompt & 61.42 & 48.22 & 52.79 & 88.83 & 59.90 & 76.65\\
ASCD & 72.59 & 54.31 & 53.81 & 87.82 & 62.94 & 71.07\\
Lora & 82.23 & 81.22 & 82.23 & 84.26 & 81.73 & 81.73\\
CAER(ours) & 70.05 & 84.77 & 86.80 & 79.19 & 81.73 & 77.66\\
\bottomrule
\end{tabular}}
\end{table}

\begin{table}[t]
\centering
\caption{Normal classification accuracy (\%) on AgriConflict.}
\label{tab:agri_normal_main}
\resizebox{\linewidth}{!}{
\begin{tabular}{lcccccc}
\toprule
Method & InternVL3 & LLaVA-OV2 & LLaVA-OV1.5 & Ministral3 & Qwen3-VL & Qwen3.5\\
\midrule
Raw & 72.59 & 89.34 & 88.32 & 57.78 & 77.66 & 68.02\\
FoV prompt & 92.39 & 94.92 & 89.85 & 64.47 & 86.80 & 85.28\\
MMMC prompt & 83.25 & 91.88 & 89.34 & 56.35 & 79.19 & 73.60\\
ASCD & 70.56 & 89.34 & 88.83 & 52.28 & 76.14 & 71.07\\
LoRA & 72.59 & 80.20 & 78.17 & 75.13 & 79.70 & 84.26\\
CAER(ours) & 72.59 & 55.84 & 53.81 & 62.44 & 64.47 & 72.59\\
\bottomrule
\end{tabular}}
\end{table}

\begin{figure}[t]
    \centering
    \includegraphics[width=\linewidth]{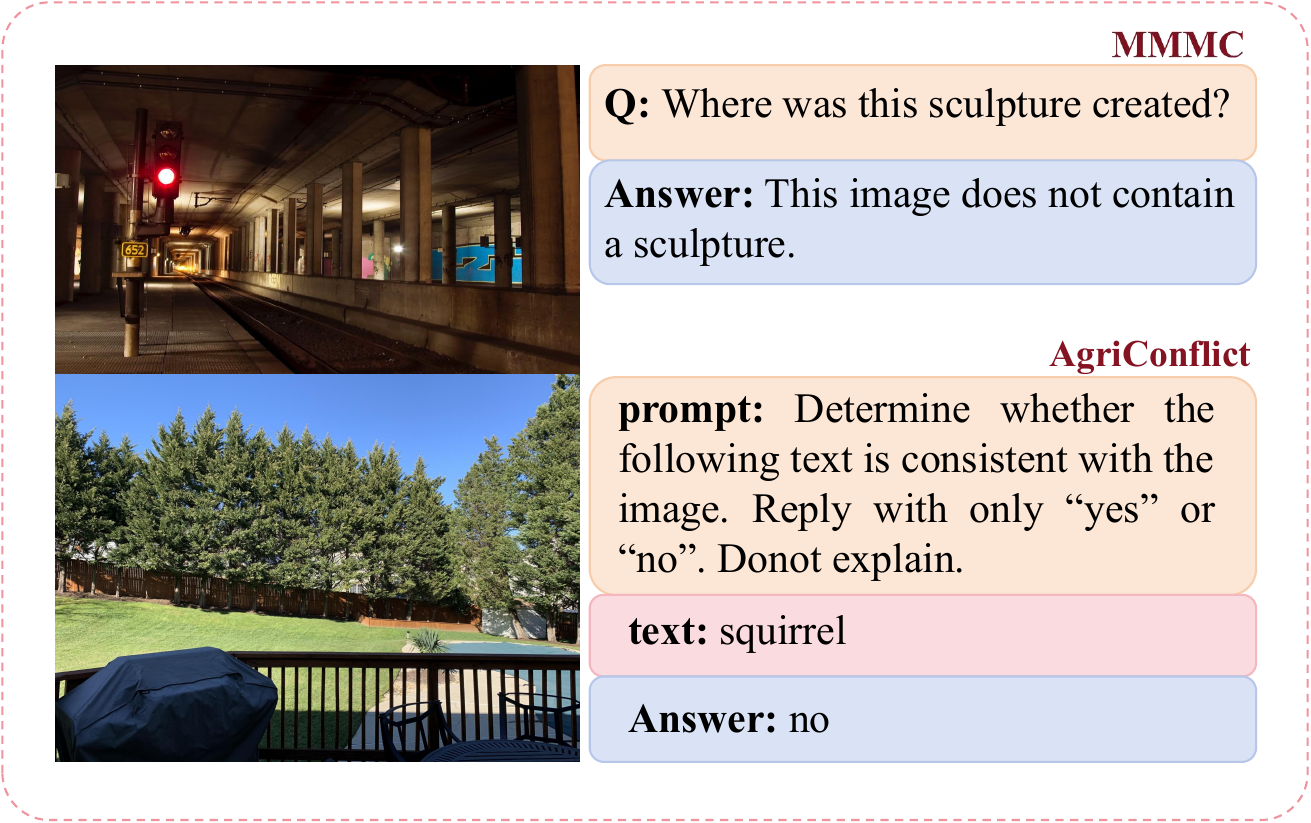}
    \caption{Examples of conflict-aware generation. CAER identifies inconsistent textual assumptions and produces responses grounded in visual evidence under both MMMC and AgriConflict scenarios.}
    \label{fig:qualitative_examples}
\end{figure}

On AgriConflict, CAER achieves competitive performance under a domain-specific image-claim consistency setting. In particular, CAER improves conflict detection accuracy over the original models on several backbones, including LLaVA-OV2-8B and LLaVA-OV1.5-8B. These results demonstrate the generalization ability of evidence routing beyond the MMMC benchmark. Figure~\ref{fig:qualitative_examples} presents qualitative examples, where CAER identifies inconsistent textual assumptions and generates responses grounded in visual evidence.

Although LoRA achieves higher conflict resolution accuracy on MMMC, it requires substantially more trainable parameters and optimization resources. Specifically, LoRA updates approximately 187--216M parameters across different backbones, while CAER introduces only 16.0--17.6M trainable parameters in Phase A and an additional 10.0--11.6M parameters for prefix experts in Phase B. Moreover, LoRA requires approximately 7--8$\times$ more training time than CAER under comparable settings. These results demonstrate that CAER provides a parameter-efficient solution for conflict-aware adaptation without modifying the underlying vision-language backbone.

We observe that CAER slightly decreases performance on some non-conflict samples, mainly due to imperfect conflict annotations. Since CAER routes inputs according to the predicted conflict status, mislabeled samples may activate inappropriate experts and affect generation. Nevertheless, CAER maintains conflict-aware improvements while preserving general multimodal capabilities.

\subsection{Ablation Study}

\paragraph{Loss Weight Analysis in Phase A.}
We conduct an ablation study on different loss configurations for training the Conflict-Aware Evidence Router in Phase A. Specifically, we vary the weights of $\mathcal{L}_{\mathrm{span}}$ and $\mathcal{L}_{\mathrm{rank}}$ while keeping $\mathcal{L}_{\mathrm{status}}$ unchanged. The results are shown in Table~\ref{tab:phasea_loss_ablation}.

\begin{table}[t]
\centering
\caption{Ablation study of different loss configurations in Phase A.}
\label{tab:phasea_loss_ablation}
\begin{tabular}{cccc}
\toprule
$\lambda_1$
& $\lambda_2$
& NCA (\%)
& CRA (\%) \\
\midrule
0.4 & 0.0 & 81.30 & 78.00 \\
0.4 & 0.1 & 82.75 & 78.15 \\
0.8 & 0.1 & \textbf{82.90} & 78.40 \\
0.8 & 0.1 (MB=8) & 82.00 & 78.60 \\
0.2 & 0.1 & 81.75 & 78.20 \\
0.8 & 0.2 & 82.35 & \textbf{78.85} \\
\bottomrule
\end{tabular}
\end{table}

We select $\lambda_{1}=0.8$ and $\lambda_{2}=0.2$ based on the highest CRA among different Phase-A loss configurations. This setting is adopted in all subsequent experiments.

\section{Conclusion}

In this work, we study vision-language conflict handling in MLLMs, where textual inputs may contain false assumptions inconsistent with visual evidence. We propose CAER, a backbone-agnostic framework that introduces a span-grounded evidence router to associate conflict-related claims with visual evidence and a dual prefix expert routing mechanism for status-conditioned generation while keeping MLLMs frozen. Extensive experiments on the MMMC benchmark and our newly constructed AgriConflict dataset demonstrate that CAER improves conflict detection and correction across open-source MLLMs with fewer trainable parameters and lower optimization costs than full parameter adaptation approaches.

\clearpage
\bibliography{references}

\end{document}